\documentclass[10pt,twocolumn,letterpaper]{article}

\usepackage[pagenumbers]{cvpr} 

\usepackage{amsmath,amsfonts,bm}

\def\eqref#1{equation~\ref{#1}}

\def\1{\bm{1}}

\def\rvd{{\mathbf{d}}}

\def\rvn{{\mathbf{n}}}

\def\rvx{{\mathbf{x}}}

\DeclareMathAlphabet{\mathsfit}{\encodingdefault}{\sfdefault}{m}{sl}
\SetMathAlphabet{\mathsfit}{bold}{\encodingdefault}{\sfdefault}{bx}{n}

\def\gA{{\mathcal{A}}}

\def\gD{{\mathcal{D}}}
\def\gE{{\mathcal{E}}}
\def\gF{{\mathcal{F}}}

\def\gL{{\mathcal{L}}}
\def\gM{{\mathcal{M}}}

\def\gT{{\mathcal{T}}}
\def\gU{{\mathcal{U}}}
\def\gV{{\mathcal{V}}}

\usepackage{multirow}
\usepackage{colortbl}
\usepackage[dvipsnames]{xcolor}

\definecolor{rred}{RGB}{245, 152, 153}
\definecolor{oorange}{RGB}{253, 205, 154}
\definecolor{yyellow}{RGB}{248,244,140}
\def\nickname{URHand}

\definecolor{cvprblue}{rgb}{0.21,0.49,0.74}
\usepackage[pagebackref,breaklinks,colorlinks,citecolor=cvprblue]{hyperref}
\usepackage{xcolor}
\usepackage{marvosym}

\title{\nickname: Universal Relightable Hands}

\author{Zhaoxi Chen$^{1, 2*}$\quad 
Gyeongsik Moon$^{1}$\quad 
Kaiwen Guo$^{1}$\quad 
Chen Cao$^{1}$ \quad 
Stanislav Pidhorskyi$^{1}$\\
Tomas Simon$^{1}$ \quad 
Rohan Joshi$^{1}$\quad 
Yuan Dong$^{1}$\quad 
Yichen Xu$^{1}$\quad 
Bernardo Pires$^{1}$\\
He Wen$^{1}$\quad 
Lucas Evans$^{1}$\quad
Bo Peng$^{1}$\quad 
Julia Buffalini$^{1}$\quad 
Autumn Trimble$^{1}$\\ 
Kevyn McPhail$^{1}$\quad 
Melissa Schoeller$^{1}$\quad 
Shoou-I Yu$^{1}$\quad 
Javier Romero$^{1}$ \\
Michael Zollhöfer$^{1}$\quad
Yaser Sheikh$^{1}$\quad  
Ziwei Liu$^{2}$\textsuperscript{\Letter} \quad 
Shunsuke Saito$^{1}$\textsuperscript{\Letter} \\
$^1$Codec Avatars Lab, Meta \qquad $^2$Nanyang Technological University \\
\url{https://frozenburning.github.io/projects/urhand}
}

\begin{document}

\twocolumn[{%
\renewcommand\twocolumn[1][]{#1}%
\vspace{-0.2in}
\maketitle
\begin{center}
\vspace{-25pt}
    \centering
    \captionsetup{type=figure}
    \includegraphics[width=0.95\textwidth]{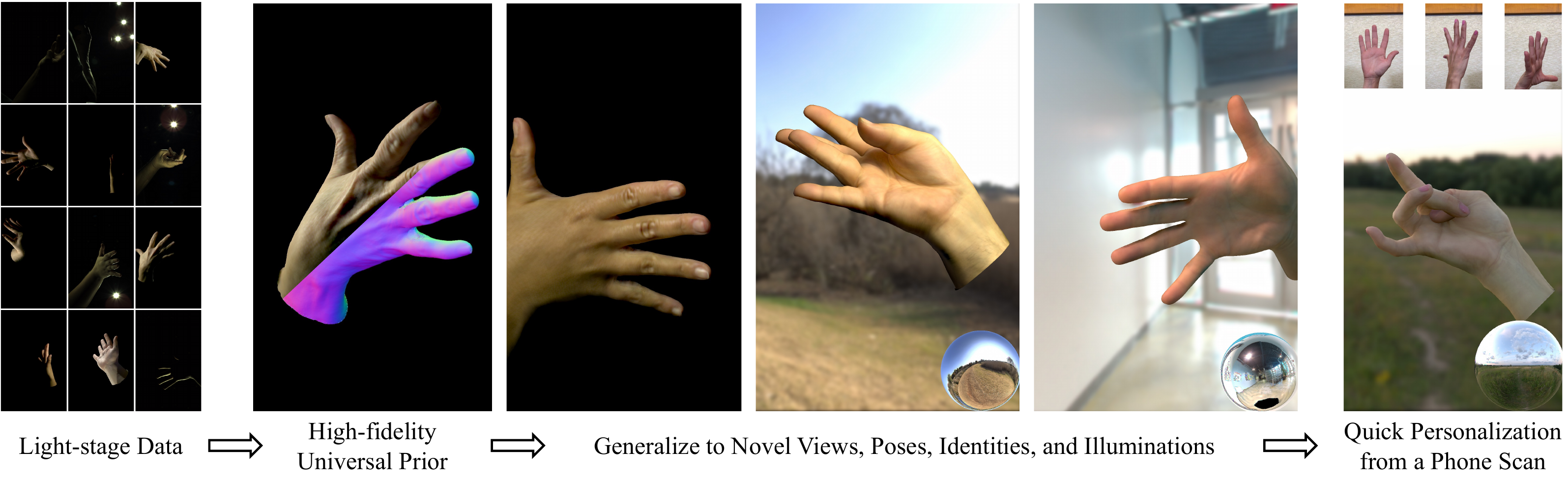}
    \vspace{-0.1in}
    \captionof{figure}{\textbf{\nickname} (\textit{a.k.a.} Your Hand). Our model is a high-fidelity \textbf{U}niversal prior for \textbf{R}elightable \textbf{Hand}s built upon light-stage data. It generalizes to novel viewpoints, poses, identities, and illuminations, which enables quick personalization from a phone scan.
    }\label{fig:teaser}
\end{center}%
}]

\let\thefootnote\relax\footnotetext{$^*$This work was done during an internship at Meta}
\let\thefootnote\relax\footnotetext{\textsuperscript{\Letter}Corresponding authors}

\begin{abstract}
\vspace{-0.15in}
Existing photorealistic relightable hand models require extensive identity-specific observations in different views, poses, and illuminations, and face challenges in generalizing to natural illuminations and novel identities. 
To bridge this gap, we present \textbf{\nickname}, the first universal relightable hand model that generalizes across viewpoints, poses, illuminations, and identities. 
Our model allows few-shot personalization using images captured with a mobile phone, and is ready to be photorealistically rendered under novel illuminations. 
To simplify the personalization process while retaining photorealism, we build a powerful universal relightable prior based on neural relighting from multi-view images of hands captured in a light stage with hundreds of identities.
The key challenge is scaling the cross-identity training while maintaining personalized fidelity and sharp details without compromising generalization under natural illuminations.
To this end, we propose a spatially varying linear lighting model as the neural renderer that takes physics-inspired shading as input feature. By removing non-linear activations and bias, our specifically designed lighting model explicitly keeps the linearity of light transport.
This enables single-stage training from light-stage data while generalizing to real-time rendering under arbitrary continuous illuminations across diverse identities.
In addition, we introduce the joint learning of a physically based model and our neural relighting model, which further improves fidelity and generalization.
Extensive experiments show that our approach achieves superior performance over existing methods in terms of both quality and generalizability.
We also demonstrate quick personalization of \nickname\ from a short phone scan of an unseen identity.
\end{abstract}
    
\vspace{-0.28in}
\section{Introduction}
\label{sec:intro}
We engage our hands for various tasks throughout the day, and they consistently remain within our field of view. This constant visibility of our hands makes them one of the most frequently seen parts of our body, playing a central role in self-embodiment. To seamlessly reproduce this experience for games or social telepresence, an ideal hand representation in a digital medium is photorealistic, personalized, and importantly, relightable for coherent appearance in any environment. Our objective is to enable the quick creation of such a hand model for any individual given lightweight input such as a phone scan, without going through an expensive capture process in a production studio (Figure~\ref{fig:teaser}).

Approaches to build a photorealistic relightable hand model can be broadly categorized into one of two philosophies. On the one hand, physically based rendering models~\cite{weyrich2006analysis,li2022nimble} can generalize to various illuminations through costly offline path-tracing, but typically lack photorealism under a real-time constraint. Additionally, accurately estimating material parameters remains a non-trivial challenge from unconstrained inputs and the quality is often bounded by the expressiveness of the physical models. 
On the other hand, neural relighting~\cite{bi2021deep,Iwase_2023} recently achieves impressive photorealism in real-time by directly inferring the outgoing radiance from illumination conditions. However, for generalization to natural illuminations to be possible these approaches require expensive data augmentation with teacher-student distillation, where the student model learns to match with offline renderings produced by the teacher model under natural illuminations. Most importantly, cross-identity generalization remains an open problem in both camps. 

In this work, we propose \nickname, the first \textbf{U}niversal \textbf{R}elightable \textbf{Hand} model that generalizes across viewpoints, motions, illuminations, and identities.
To achieve the best trade-off between generalization and fidelity, our work exploits both physically based rendering and data-driven appearance modeling from neural relighting. 
More specifically, we incorporate known physics, such as the linearity of light transport~\cite{debevec2000acquiring} and surface reflections in the inductive bias of the neural relighting framework. 
We modulate non-linear layers conditioned by pose and identity with linear layers conditioned by spatially varying physically based shading~\cite{disneybrdf}. 
This explicitly ensures linearity between input lighting features and output radiance. Thus, it enables environment map relighting without an expensive two-stage teacher-student distillation process commonly used in existing models~\cite{bi2021deep, Iwase_2023}. 
Our single-stage training enabled by linearity preservation makes cross-identity training more scalable with better generalization to novel illuminations.

Furthermore, we observe that the quality of the input shading features directly influences both generalization and fidelity of the final neural relighting outputs. 
Inspired by recent inverse rendering techniques~\cite{zhang2021physg,zhang2021nerfactor,chen2022relighting4d}, we introduce an additional physical branch that estimates the material parameters and high-resolution geometry via inverse rendering, from which we produce the input lighting features to the neural branch.
The physical branch prevents the neural branch from overfitting by reducing hallucinations, and the neural branch compensates for the complex global light transport effects, such as subsurface scattering, that cannot be well captured by the physical branch.
In addition, the proposed physics-based refinement improves the accuracy of the tracking geometry with fine details such as wrinkles.
Combining it with our novel lighting-aware adversarial loss, our method achieves highly detailed relighting with various illuminations under any pose for novel identities. 

We run an extensive ablation study as well as comparisons with baseline methods. The experiments demonstrate the efficacy of our hybrid neural-physical relighting method by outperforming other methods quantitatively and qualitatively. We also demonstrate the quick personalization of URHand from a phone scan, and relighting with arbitrary natural illuminations. In summary, our contributions are:
\begin{itemize}
    \item The first method to learn a universal relightable hand model that generalizes to novel views, poses, illuminations, and identities.

    \item A spatially varying linear lighting model that generalizes to continuous illuminations without expensive distillation, enabling high-fidelity neural rendering and scalable training with multiple identities.

    \item A hybrid neural-physical relighting framework that leverages the best of both approaches to achieve high fidelity and generalization at the same time.

    \item The quick personalization of our universal prior to create a photorealistic and relightable hand from a phone scan.
\end{itemize}

\section{Related Work}
\label{sec:review}
\noindent \textbf{3D Hand Modeling.} 
Human hand modeling is an active research field within vision and graphics. Early works focus more on 3D geometry and representation including mixture of 3D Gaussians~\cite{sridhar2015fast, sridhar2013interactive}, sphere meshes~\cite{tkach2016sphere}, and triangular meshes~\cite{romero2017mano, ballan2012motion, de2011model}. These parametric hand models facilitate 3D hand pose estimation from 2D observations~\cite{moon2018v2v, moon2020i2l, mueller2017real, mueller2018ganerated, mueller2019real}. Recent works also incorporate physical priors~\cite{li2022nimble, zheng2022simulation, wang2019hand, smith2020constraining, moon2020deephandmesh} to model more accurate non-rigid deformation and articulation of the hand geometry. The recent advances of neural fields~\cite{xie2022neural, tewari2022advances} also enable the learning of personalized articulated models~\cite{karunratanakul2021skeleton}. Beyond geometry modeling, achieving a lifelike appearance for hands~\cite{qian2020html} is paramount for realistic rendering and animation. Handy~\cite{potamias2023handy} learns a texture space by using generative adversarial networks for better photorealism and generalization. More recently, some methods~\cite{corona2022lisa, chen2023hand, mundra2023livehand} showcase the modeling of animatable hands from monocular/multiview captures. However, the appearance models of these approaches simply bake the captured illumination and cannot be rendered under novel illuminations. NIMBLE~\cite{li2022nimble} builds PCA reflectance maps including diffuse, normal, and specular maps from light-stage data. DART~\cite{gao2022dart} also supports accessories. However, physically based materials are expensive to render with global illumination which often limits their rendering fidelity with a real-time constraint. RelightableHands~\cite{Iwase_2023} enables the photorealistic relighting of hands in real-time using a neural appearance model. However, the method only supports person-specific modeling and generalization to unseen identity is not possible. In contrast, our approach generalizes across poses, illuminations, and identities, supporting relightable hand modeling of unseen identities from a phone scan.

\noindent \textbf{Image-based Relighting.}
Given the linearity of light transport, Debevec \etal~\cite{debevec2000acquiring} propose to render human faces under novel illuminations by linear combinations of sampled reflectance fields from one-light-at-a-time (OLAT) captures. A series of follow-up work~\cite{meka2019deep, xu2018deep, wenger2005performance, sun2020light} enable dynamic relighting and capture through learning-based approaches. Meanwhile, another line of work aims at intrinsic decomposition, enabling physically based rendering with disentangled geometry and reflectance in the image space~\cite{sengupta2018sfsnet, nestmeyer2020learning, lagunas2021single, kanamori2019relighting, hou2021towards, hou2022face}. Recent advances in neural rendering~\cite{ji2022geometry, totalrelighting, lumos} learn to match with ground truth using non-linear neural networks given the lighting features from a simple physical shading model. Yet, image-based relighting suffers from 3D inconsistency and flickering due to the lack of an underlying 3D representation.

\noindent \textbf{Model-based Relighting.}
To address the lack of 3D consistency in image-based relighting, one can leverage a shared 2D parameterization~\cite{yamaguchi2018high, zhang2021neural} or template models~\cite{bi2021deep, Iwase_2023, karunratanakul2023harp, chen2022relighting4d} for model-based relighting. 
Most model-based relighting approaches~\cite{chen2022relighting4d, jin2023tensoir, iqbal2023rana, qiu2023relitalk, wang2023sunstage} rely on the intrinsic decomposition of geometry and reflectance followed by a physically based appearance model. While they achieve generalization to novel conditions, the fidelity is typically limited due to the lack of expressiveness in the underlying parametric BRDFs. On the other hand, methods with neural renderers support complex global illumination effects learned from captured data under point lights. However, generalization to continuous environments requires the expensive teacher-student training framework~\cite{bi2021deep, Iwase_2023}, which is difficult to scale to cross-identity training. On the contrary, our spatially varying linear network achieves generalization to any type of illumination without additional training. Concurrently, Yang \etal~\cite{yang2023towards} propose a linear lighting model for face relighting that eliminates the need of teacher-student distillation. While the motivation is similar, we observe that their holistic light representation does not generalize well for hands due to drastic visibility change by articulation (see Sec.~\ref{sec:ablation} for analysis). 
\section{Preliminary}
\label{sec:preliminary}
\noindent \textbf{Data Acquisition.}
\label{sec:data-ac}
We use a multiview capture system consisting of 150 cameras and 350 LED lights to capture dynamic hands with time-multiplexed illuminations by interleaving fully lit (all lights on) and partially lit every other frame. Instead of OLAT, our partially lit frames use $L=5$ grouped lights to increase brightness and reduce motion blur as in~\cite{bi2021deep,li2023megane,Iwase_2023}. Images are captured in the resolution of $4096 \times 2668$ at 90 fps. We first reconstruct per-frame 3D meshes using~\cite{guo2019relightables} and detect 3D hand keypoints using~\cite{li2019rethinking} followed by triangulation from fully lit frames.

For partially lit frames, we leverage spherical linear interpolation over the pose parameter of adjacent fully lit frames to obtain the hand pose of partially lit frames. Our dataset contains 93 different identities in diverse hand motions with an average of 42000 frames for each identity.

\noindent \textbf{Linearity of Light Transport.}
\label{sec:linearity}
To render subjects in arbitrary illuminations, natural illumination is treated as a linear combination of distant point lights given the linearity of light transport~\cite{debevec2000acquiring}. 
Specifically, given the appearance value $\mathbf{C}^i$ based on the $i$-th point light, the final color $\mathbf{C}$ is computed as a linear combination of all light sources, \ie $\mathbf{C} = \sum^{L}_{i=1}b_i \mathbf{C}^i$, where $L$ denotes the number of lights, and $b_i$ is the intensity of each light. Given a light transport function $f(b) = \mathbf{C}$, we define it as linear \wrt $b$ such that:
\begin{equation}
\label{eq:linearity}
\small
    f(b) = f(\sum^{L}_{i=1} b_i) = \sum^{L}_{i=1} f(b_i), \;\; b = \sum^{L}_i{b_i}.
\end{equation}

\noindent \textbf{Hand Geometry Modeling.} 
\label{sec:uhm}
Similar to~\cite{moon2020deephandmesh}, our 3D hand representation is based on a mesh template with vertex offsets predicted by a neural network. The 3D hand can be driven by linear blend skinning (LBS) and represent identity- and pose-specific deformations.

Specifically, we design an autoencoder to obtain accurate hand tracking and geometry. 
The encoder learns to predict identity-dependent latent codes and 3D hand poses from the input fully lit frames.
Given an articulated generic mesh template $\Bar{\gM} = \{\gV, \gF, \gU, \theta\}$ with vertices $\gV \in \mathbb{R}^{n_{\gV} \times 3}$, faces $\gF \in \mathbb{R}^{n_{\gF} \times 3}$, texture coordinates $\gU \in \mathbb{R}^{n_{\gV}\times2}$, and 3D pose $\theta \in \mathbb{R}^{60\times3}$ in Euler angles and latent codes $z$, the decoder learns to predict the 3D offset of all vertices as $\delta \gV \in \mathbb{R}^{n_{\gV}\times3}$. 
We use $n_{\gV} = 15930$ and $n_{\gF} = 32340$.
The tracked hand mesh will be represented as $\gM = \{\gV + \delta \gV, \gF, \gU, \theta\}$. 
Please refer to the supplementary material for more details.
\begin{figure*}
    \centering
    \vspace{-0.3in}
    \includegraphics[width=0.9\linewidth]{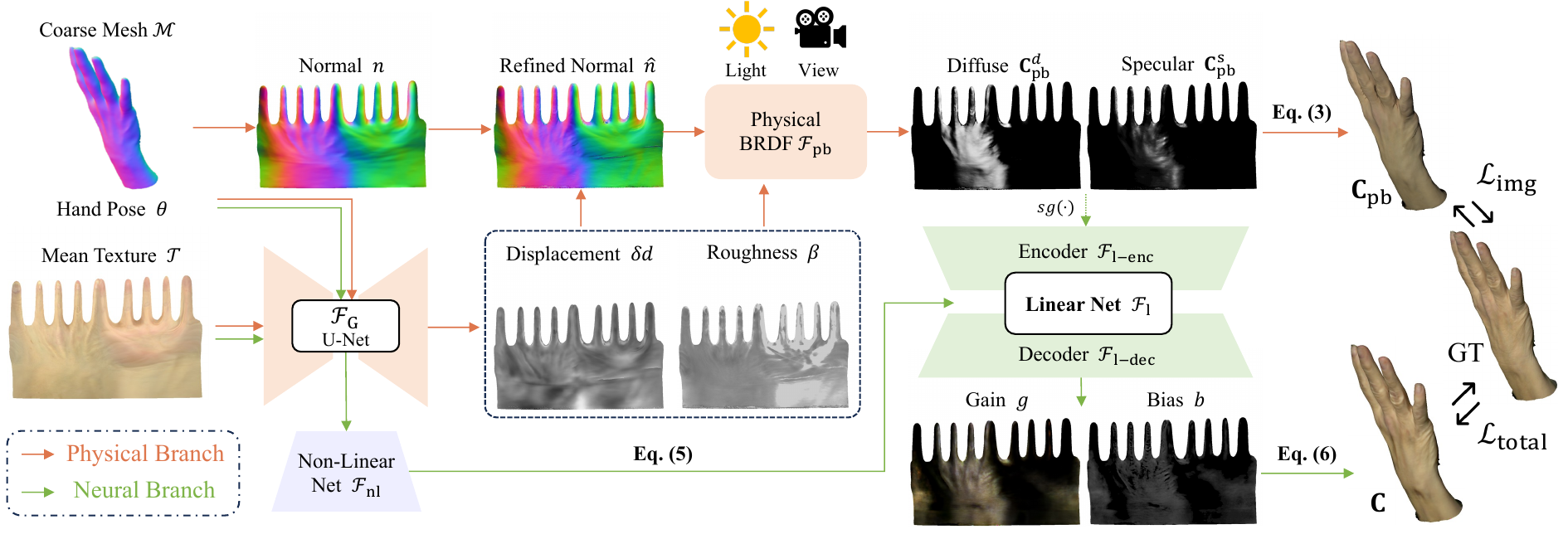}
    \vspace{-0.1in}
    \caption{\textbf{Overview of \nickname.} Our model takes as input a mean texture $\gT$, hand pose $\theta$, and a coarse mesh $\gM$ for each identity. The physical branch (Sec.~\ref{sec:phys-refiner}) focuses on geometry refinement and providing accurate shading features for the neural branch (Sec.~\ref{sec:linear-model}). The core of the neural branch is the linear lighting model which takes as input the physics-inspired shading features from the physical branch. The neural branch learns to predict the gain and bias map over the mean texture. We leverage a differentiable rasterizer for rendering and minimize the loss of both branches against ground truth images (Sec.~\ref{sec:training}). The $sg(\cdot)$ denotes the stop-gradient operation.}
    \label{fig:overview}
    \vspace{-0.1in}
\end{figure*}

\section{Universal Relightable Hands}
\label{sec:method}
Our goal is to build a universal relightable shape and appearance model for human hands that can be rendered for any identity under arbitrary illumination in real-time. To this end, we learn a relightable appearance model from cross-identity light-stage captures based on grouped point lights. 

In this section, we will introduce our learning framework, \nickname, which learns to relight target hands in different poses and views. The core of our model is a spatially varying linear lighting model that preserves the linearity of light transport, which enables the generalization to arbitrary illuminations by training with monochrome group lights only. 
Our model consists of two parallel rendering branches,  physical and neural. The physical branch (Sec.~\ref{sec:phys-refiner}) focuses on refining geometry and providing accurate shading features as an illumination proxy for the neural branch. The neural branch (Sec.~\ref{sec:linear-model}) learns the final appearance of hands with global illumination. These two branches are trained jointly in an end-to-end manner with our tailored loss functions (Sec.~\ref{sec:training}). Finally, we use this universal prior to quickly personalize a relightable hand model from few-shot observations (Sec.~\ref{sec:instant}).

\subsection{Physically based Geometry Refinement}
\label{sec:phys-refiner}

The physical branch employs online physically based rendering to render images using a parametric BRDF.
We optimize the material parameters of the BRDF via inverse rendering. The goal of the physical branch is two-fold: 1) further refine the initial hand geometry for better alignment, and 2) provide generalizable lighting features that best approximate the specular reflection and diffuse shading to prevent overfitting in the neural relighting.

We illustrate the physical branch in Fig.~\ref{fig:overview}. The physical branch estimates a parametric Disney BRDF~\cite{disneybrdf} $\mathcal{F}_{\mathrm{pb}}$. We use a 2D U-Net $\mathcal{F}_{\mathrm{G}}$ to infer a displacement map $\delta d \in \mathbb{R}^{1024\times1024\times3}$ and a roughness map $\beta \in \mathbb{R}^{1024\times1024}$ in UV space. Instead of predicting a normal map directly, we use the predicted displacement map to update the normal map on top of the unwrapped coarse mesh in the UV space, which makes it easier to infer high-frequency geometric details on a smooth base surface~\cite{yifan2022geometryconsistent}.
To support cross-identity modeling with pose-dependent geometry and appearance change, $\mathcal{F}_{\mathrm{G}}$ takes as input the hand pose $\theta$ and the unwrapped mean texture $\mathcal{T} \in \mathbb{R}^{1024\times1024\times3}$ for each identity, where the hand pose is concatenated to the bottleneck representation of the U-Net. 

Specifically, the refined surface $\hat{\rvx}$ is obtained by adding the displacement $\delta d$ to the base surface $\rvx$ derived from the coarse mesh $\mathcal{M}$ along the direction of the normal $\rvn$:
\begin{equation}
\label{eq:displacement}
    \hat{\rvx} = \rvx + \delta d \cdot \rvn,
\end{equation}
where positional map $\hat{\rvx}$ is then used to obtain the refined normal $\hat{\rvn}$ in the UV space for physically based rendering. In our implementation, we apply a sigmoid activation followed by a scaling factor of 3 in order to constrain the range of displacement to $\pm 3 \mathrm{mm}$.

The refined normal $\hat{\rvn}$ and roughness $\beta$ are fed into $\gF_{\mathrm{pb}}$ that considers only the first bounce. Given an illumination $\mathcal{L}$, the final color $\mathbf{C}_{\mathrm{pb}}$ from camera view $\rvd$ is computed as:

\begin{equation}
\label{eq:pbr}
    \mathbf{C}_{\mathrm{pb}}(\hat{\rvx}, \rvd, \mathcal{L}) =
    \int L_i(\mathcal{L}, \hat{\rvx}, \mathbf{\omega}_i) \gF_{\mathrm{pb}}(\hat{\rvx}, \mathbf{\omega}_i, \rvd, \beta)(\mathbf{\omega}_i \cdot \hat{\rvn}) d \mathbf{\omega}_i,
\end{equation}
where $L_i(\hat{\rvx}, \mathbf{\omega}_i)$ is the incident light from direction $\mathbf{\omega}_i$. The physically rendered texture map $\mathbf{C}_{\mathrm{pb}}$ can be decomposed into physically based shading feature, \ie $\mathbf{F}_{\mathrm{pb}}(\mathcal{L}) = \{\mathbf{C}^d_{\mathrm{pb}}, \mathbf{C}^s_{\mathrm{pb}}\}$, according to the equation $\mathbf{C}_{\mathrm{pb}} = \mathbf{C}_{\mathrm{pb}}^d \odot \gT + \mathbf{C}_{\mathrm{pb}}^s$, where $\odot$ is the element-wise multiplication, and $\gT$ is the mean texture approximating albedo. 
Note that, this computation is directly performed in the UV space. 

\subsection{Linear Lighting Model}
\label{sec:linear-model}
The rendering based on a parametric BRDF generalizes to novel illuminations, but they lack correct global light transport effects such as subsurface scattering. To enable relighting with global illumination in real-time, we introduce a neural renderer $\gF_{\mathrm{R}}$. Given a target illumination, the outgoing radiance is computed as:
$
    \mathbf{C} = \gF_{\mathrm{R}}(\gM, \gT, \mathbf{F}_{\mathrm{pb}}(\mathcal{L}), \theta).
$

Our key insight is that removing the non-linear activation layers and bias in a convolutional neural network preserves the linearity with respect to the input features. Since our input feature is the physically based shading  $\mathbf{F}_{\mathrm{pb}}(\mathcal{L})$, our network satisfies the following: 
\begin{equation}
\label{eq:linear-model}
\footnotesize
    \sum^L_{i=1} \gF_{\mathrm{R}}(\gM, \gT, \mathbf{F}_{\mathrm{pb}}(\mathcal{L}_i), \theta) = \gF_{\mathrm{R}}(\gM, \gT, \sum_{i=1}^{L}{\mathbf{F}_{\mathrm{pb}}(\mathcal{L}_i)}, \theta).
\end{equation}
Since the physically based rendering in Eq.~\ref{eq:pbr} is energy preserving, the network holds the linearity \wrt the input illumination. Therefore, our network can produce accurate relighting with continuous environment maps by training only with discrete point lights without additional distillation.

We illustrate the architecture of $\gF_{\mathrm{R}} = \{\gF_{\mathrm{l}}, \gF_{\mathrm{nl}}\}$ in Fig.~\ref{fig:overview}. It consists of a linear ($\gF_{\mathrm{l}}$) and a non-linear ($\gF_{\mathrm{nl}}$) branch, where the linear branch consists of an encoder $\gF_{\mathrm{l-enc}}$ and a decoder $\gF_{\mathrm{l-dec}}$ with physically based shading features $\mathbf{F}_{\mathrm{pb}}(\mathcal{L})$ as input. The pose- and identity-dependent features derived from the mean texture $\gT$ and pose $\theta$ are fed into a non-linear branch. We fuse the linear and non-linear feature maps in the decoder of the linear branch as follows:
\begin{equation}
\label{eq:gb-injection}
    \gF_{\mathrm{l-dec}}^{j+1} = \frac{1}{\sqrt{2}} \cdot  \mathrm{ConvT}(\gF_{\mathrm{l-enc}}^j + \gF_{\mathrm{l-dec}}^j) \odot \gF_{\mathrm{nl}}^j,
\end{equation}
where $j$ is the index of the layer, $\mathrm{ConvT}$ is the transposed convolutional layer without bias. This fusion mechanism keeps the linearity of the output \wrt the input lighting features while incorporating non-linearity \wrt identity and pose conditions.
Instead of predicting the final texture, our neural renderer predicts the texel-aligned gain map $g$ and bias map $b$, which contribute to the final texture as follows:
\begin{equation}
\label{eq:gb-rendering}
    \mathbf{C} = g \odot \gT + b \cdot \sigma_{\gT},
\end{equation}
where $\sigma_{\gT} = 64$ is the standard deviation of textures.

Important to note that since our input feature $\mathbf{F}_{\mathrm{pb}}$ is spatially varying in a texel-aligned manner similar to~\cite{totalrelighting, Iwase_2023}, we can accurately incorporate shadow information for better generalization with diverse poses.
While the concurrent work~\cite{yang2023towards} also proposes a linear lighting model using a holistic illumination representation by simply reshaping environment maps, we observe that the holistic illumination representation cannot generalize for hands due to the infinite shadow variations caused by articulation. 
The same observation is also reported in person-specific relightable hand modeling~\cite{Iwase_2023}. Please refer to Sec.~\ref{sec:experiments} for the analysis.

\subsection{Training Objectives}
\label{sec:training}
Our model is trained on multiview partially lit images in different identities and poses. The training objective $\gL_{\mathrm{total}}$ consists of three parts: reconstruction loss $\gL_{\mathrm{img}}$, lighting-aware adversarial loss $\gL_{\mathrm{GAN}}$, and L1 regularization $\gL_{\mathrm{reg}}$:
\begin{equation}
\label{eq:full-loss}
    \gL_{\mathrm{total}} = \lambda_{\mathrm{img}} \gL_{\mathrm{img}} + \lambda_{\mathrm{GAN}} \gL_{\mathrm{GAN}} + \lambda_{\mathrm{reg}} \gL_{\mathrm{reg}},
\end{equation}
where $\lambda_{*}$ are corresponding loss weights.

\noindent \textbf{Reconstruction Loss.} We leverage a sum of L1 loss $\gL_{\mathrm{MAE}}$ and perceptual loss $\gL_{\mathrm{eff}}$ based on the EfficientNet~\cite{tan2019efficientnet} backbone, \ie $\gL_{\mathrm{img}} = \gL_{\mathrm{MAE}} + \gL_{\mathrm{eff}}$. Both the renderings from the neural and the physical branch are supervised by the reconstruction loss against the ground truth images.

\noindent \textbf{Lighting-aware Adversarial Loss.} To improve the visual quality, we propose to use an adversarial loss on top of the reconstruction loss. We found that a naive image-conditioned discriminator performs poorly due to the significant appearance change in partially lit frames. To address this, we leverage a lighting-aware discriminator on multiple scales of renderings. Specifically, the discriminator $\gF_{\mathrm{D}}$ is conditioned on the diffuse and specular feature, $\{A, S\}$, which prompts the network to discriminate real and fake images given the illumination information. These lighting features are based on simple Phong reflectance~\cite{Iwase_2023} to ensure a consistent lighting prompt during training. We choose a hinge loss~\cite{lim2017geometric} operated on multi-resolution discriminated patches as the adversarial target:
\begin{equation}
\label{eq:gan-loss}
    \gL_{\mathrm{GAN}} = \log \gF_{\mathrm{D}}(I | A, S) + \log [1-\gF_{\mathrm{D}}(\hat{I} | A, S)],
\end{equation}
where $I$ is the ground truth, and $\hat{I}$ is the rendered image.

\begin{table*}[t]
    \centering
    \vspace{-0.1in}
    \begin{footnotesize}
    \caption{\textbf{Quantitative comparisons on sequences with grouped lights}.  We evaluate our method for both per-subject optimization and novel identity generalization against the state-of-the-art methods in model-based hand relighting. $^{\dagger}$Methods are evaluated on the training identity with unseen segments. $^{*}$Methods are evaluated on unseen identity during training. The top three techniques are highlighted in \textcolor{rred}{red}, \textcolor{oorange}{orange}, and \textcolor{yyellow}{yellow}, respectively. 
    }
    \label{tab:comparison}
    \setlength{\tabcolsep}{2mm}{
    \renewcommand\arraystretch{0.975}
    \resizebox{2.0\columnwidth}{!}{
    \begin{tabular}{c|ccccccccc} 
    \toprule
    \multirow{2}{*}{Method}& \multicolumn{3}{c}{\textbf{Subject 1}}  & \multicolumn{3}{c}{\textbf{Subject 2}}  & \multicolumn{3}{c}{\textbf{Subject 3}}\\ 
    & \footnotesize{PSNR\;$\uparrow$} & \footnotesize{SSIM\;$\uparrow$} & \footnotesize{LPIPS\;$\downarrow$} & \footnotesize{PSNR\;$\uparrow$} & \footnotesize{SSIM\;$\uparrow$} & \footnotesize{LPIPS\;$\downarrow$} &  \footnotesize{PSNR\;$\uparrow$} & \footnotesize{SSIM\;$\uparrow$} & \footnotesize{LPIPS\;$\downarrow$}\\  
    \midrule
    $^{\dagger}$RelightableHands~\cite{Iwase_2023}&\cellcolor{yyellow}25.97&\cellcolor{oorange}0.9301&\cellcolor{yyellow}0.1425&\cellcolor{yyellow}25.92&\cellcolor{oorange}0.9372&\cellcolor{yyellow}0.1426&\cellcolor{yyellow}27.16&\cellcolor{yyellow}0.9419&\cellcolor{yyellow}0.1280\\
    $^{\dagger}$Ours (Physical only)&23.44&0.9062&0.1708&23.25&0.9154&0.1715&24.90&0.9216&0.1510\\
    $^{\dagger}$Ours (Full model)&\cellcolor{rred}27.77&\cellcolor{rred}0.9400&\cellcolor{rred}0.1204&\cellcolor{rred}26.36&\cellcolor{rred}0.9384&\cellcolor{oorange}0.1344&\cellcolor{rred}27.75&\cellcolor{rred}0.9445&\cellcolor{oorange}0.1226\\
    \midrule
    $^{*}$Handy~\cite{potamias2023handy} + Phong&16.65&0.8235&0.3026&16.39&0.8328&0.3019&17.70&0.8311&0.2872\\
    $^{*}$Handy~\cite{potamias2023handy} + GGX&16.60&0.8212&0.3125&16.33&0.8300&0.3107&17.58&0.8288&0.2978\\
    $^{*}$Ours (Full model)&\cellcolor{oorange}26.94&\cellcolor{yyellow}0.9271&\cellcolor{oorange}0.1335&\cellcolor{oorange}26.24&\cellcolor{yyellow}0.9368&\cellcolor{rred}0.1341&\cellcolor{oorange}27.58&\cellcolor{oorange}0.9436&\cellcolor{rred}0.1197\\
    \bottomrule
    \end{tabular}}}
    \end{footnotesize}
\end{table*}

\begin{figure*}
    \centering
    \vspace{-0.1in}
    \includegraphics[width=1.0\linewidth]{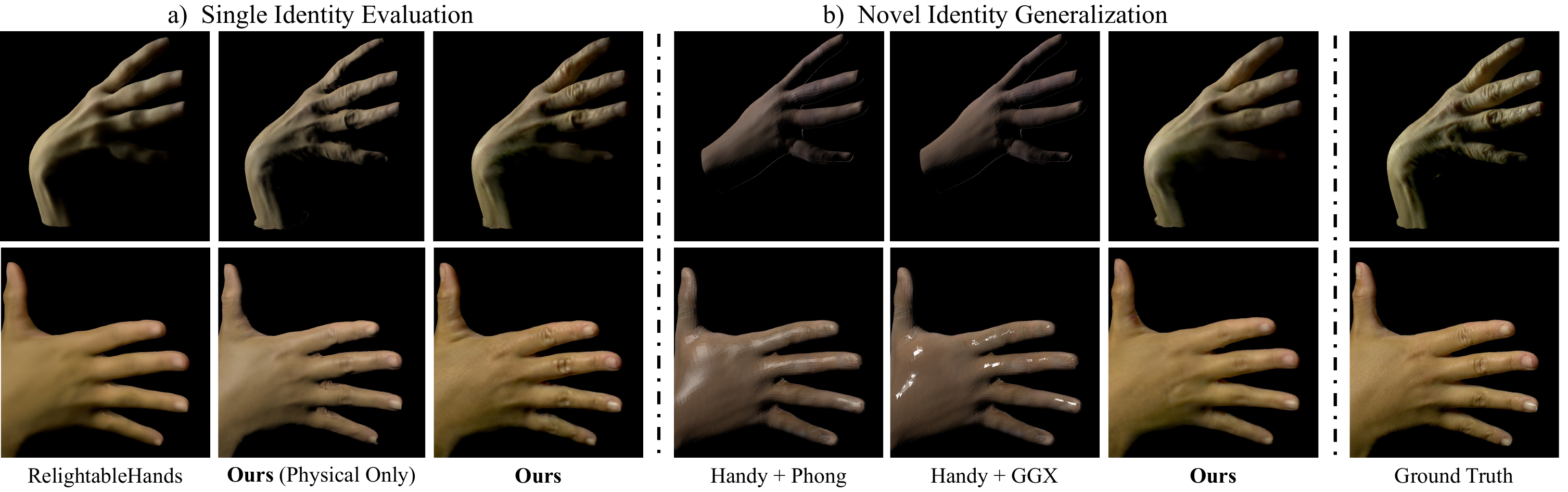}
    \caption{\textbf{Qualitative comparisons on sequences with grouped lights.} We evaluate our method for both per-subject optimization and novel identity generalization against comparison methods. a) All methods are evaluated on the training identity with unseen segments. b) Methods are evaluated on unseen identity during training. }
    \label{fig:olat-comparisons}
    \vspace{-0.15in}
\end{figure*}

\noindent \textbf{L1 Regularization on Linear Model.} We also discovered that without regularization, a linear lighting model often produces noticeable flickering. As a linear convolutional network is capacity-limited, it tends to have high variance in intermediate features, resulting in poor generalization to novel illuminations. Thus, we penalize the L1 norm of the intermediate features from all layers in the linear branch:
\begin{equation}
\label{eq:r1-reg}
    \gL_{\mathrm{reg}} = \sum^{N}_{j=1}||F^j_{\mathrm{l-enc}}||_1.
\end{equation}

\noindent \textbf{Implementation Detail.}
The optimizable modules are $\{\gF_{\mathrm{G}}, \gF_{\mathrm{R}}, \gF_{\mathrm{D}}\}$. We use the Adam~\cite{kingma2014adam} optimizer and set the loss weights as $\lambda_{\mathrm{img}} = 1.0$, $\lambda_{\mathrm{GAN}} = 0.01$, and $\lambda_{\mathrm{reg}} = 0.01$, respectively. We train the model for 2M iterations distributed on 8 A100 GPUs with a batch size of 24 in total. The initial learning rate is $1\times 10 ^ {-4}$ which decays to $3\times 10^{-5}$ with a multistep learning rate scheduler. We describe the detailed network architecture and more training details in the supplementary.

\noindent \textbf{Runtime Analysis.} Our proposed linear lighting model is not only scalable and generalizable but also efficient for real-time rendering. Our model achieves 38 FPS (25.7 ms) given grouped lights as input and 31 FPS (31.9 ms) given environmental maps as input on an NVIDIA A100 GPU.

\section{Experiments}
\label{sec:experiments}

\subsection{Evaluation Protocols}
To quantitatively evaluate the fidelity of each method, we use Peak Signal-to-Noise Ratio (PSNR), Structural Similarity Index Measure~\cite{ssim} (SSIM), and Learned Perceptual Image Patch Similarity~\cite{zhang2018perceptual} (LPIPS) as metrics. To solely evaluate the quality of rendered hands, we only take the foreground according to the mask obtained from the refined hand geometry. We exclude several segments from training data to evaluate the generalization of our model to novel poses. All metrics are evaluated on 1,000 images randomly sampled from those held-out segments.

For fair comparisons with existing works, we train and evaluate all methods in the single-identity setting except mentioned. For ablation studies, we train with ten identities and test on 1) unseen segments from a train subject, 2) unseen segments from an unseen subject, and 3) an unseen illumination from a train subject, which orthogonally evaluate the generalization of our model to novel poses, identities, and illuminations.

\begin{table*}[t]
    \centering
    \begin{footnotesize}
    \vspace{-0.1in}
    \caption{\textbf{Ablation studies of the linear lighting model on sequences with grouped lights}. The ``Non-linear'' denotes the model that does not satisfy the linearity of light transport while the ``Linearity Consistency'' denotes the non-linear model with regularizations to constrain the linearity between output and input. Moreover, in contrast to MLP-based linear model~\cite{yang2023towards}, our model is a spatially varying linear model. The top three techniques are highlighted in \textcolor{rred}{red}, \textcolor{oorange}{orange}, and \textcolor{yyellow}{yellow}, respectively.}
    \label{tab:abl-llm}
    \setlength{\tabcolsep}{2mm}{
    \renewcommand\arraystretch{0.975}
    \resizebox{2.0\columnwidth}{!}{
    \begin{tabular}{c|ccccccccc} 
    \toprule
    \multirow{2}{*}{Method}& \multicolumn{3}{c}{\textbf{Trained Subject}}  & \multicolumn{3}{c}{\textbf{Unseen Subject}}&\multicolumn{3}{c}{\textbf{Fully lit}}\\ 
    & \footnotesize{PSNR\;$\uparrow$} & \footnotesize{SSIM\;$\uparrow$} & \footnotesize{LPIPS\;$\downarrow$} & \footnotesize{PSNR\;$\uparrow$} & \footnotesize{SSIM\;$\uparrow$} & \footnotesize{LPIPS\;$\downarrow$}& \footnotesize{PSNR\;$\uparrow$} & \footnotesize{SSIM\;$\uparrow$} & \footnotesize{LPIPS\;$\downarrow$} \\  
    \midrule
    Non-linear&\cellcolor{oorange}25.79&\cellcolor{rred}0.9283&\cellcolor{oorange}0.1364&\cellcolor{oorange}23.89&\cellcolor{rred}0.9145&\cellcolor{oorange}0.1523&\cellcolor{yyellow}17.50&\cellcolor{oorange}0.9070&\cellcolor{yyellow}0.1717\\
    Linearity Consistency&\cellcolor{yyellow}24.77&\cellcolor{yyellow}0.9122&\cellcolor{yyellow}0.1560&\cellcolor{yyellow}23.11&\cellcolor{oorange}0.9124&\cellcolor{yyellow}0.1537&\cellcolor{oorange}19.75&\cellcolor{yyellow}0.9013&\cellcolor{oorange}0.1619\\
    MLP-based Linear~\cite{yang2023towards}&22.19&0.8727&0.1885&21.56&0.8787&0.1823&11.56&0.8644&0.2259\\
    \textbf{Ours} (Linear Model)&\cellcolor{rred}26.01&\cellcolor{oorange}0.9270&\cellcolor{rred}0.1336&\cellcolor{rred}24.70&\cellcolor{yyellow}0.9121&\cellcolor{rred}0.1520&\cellcolor{rred}20.61&\cellcolor{rred}0.9153&\cellcolor{rred}0.1576\\
    \bottomrule
    \end{tabular}}}
    \end{footnotesize}
    \vspace{-0.1in}
\end{table*}

\subsection{Comparisons}

We compare our approach with the state-of-the-art 3D hand relighting and reconstruction methods. RelightableHands~\cite{Iwase_2023} reconstructs relightable appearance via per-identity optimization. 
Handy~\cite{potamias2023handy} predicts the UV texture of the hand from in-the-wild images. For fair comparisons, we apply physically based renderers on top of the predicted texture to enable relightable appearances. 
We also compare with our physical branch using the same loss functions. This provides a fair comparison with physically based relighting.
For this reason, we omit the comparison with other existing physically based relightable hand methods such as HARP~\cite{karunratanakul2023harp} and NIMBLE~\cite{li2022nimble}.
Please refer to the supplementary for our detailed implementations of these methods.

We present quantitative results in Table~\ref{tab:comparison}. For per-identity training, our method significantly outperforms baseline methods on all metrics, which highlights the effectiveness of our key designs. As shown in Figure~\ref{fig:olat-comparisons}, our method is able to reproduce detailed geometry, \eg wrinkles and nails, together with high-fidelity details like specularities and shadows. Handy~\cite{potamias2023handy} fails to reproduce the base texture of the target hand due to their data-driven latent texture space. Its combinations of simple physical renderers lead to artifacts of the relit results. RelightableHands~\cite{Iwase_2023} can reproduce correct shading and appearance. However, the quality of geometric details and specularity is not on par with the proposed method, which demonstrates the effectiveness of our hybrid neural-physical approach.
Furthermore, the generalizability of our method is showcased in the right side of Figure~\ref{fig:olat-comparisons}, where the shown test subjects are withheld from the training set.
Although the quality slightly degrades compared with per-identity optimization, it still outperforms all other baselines by a large margin.

\subsection{Ablation Studies}
\label{sec:ablation}

\begin{figure}
    \centering
    \includegraphics[width=1.0\linewidth]{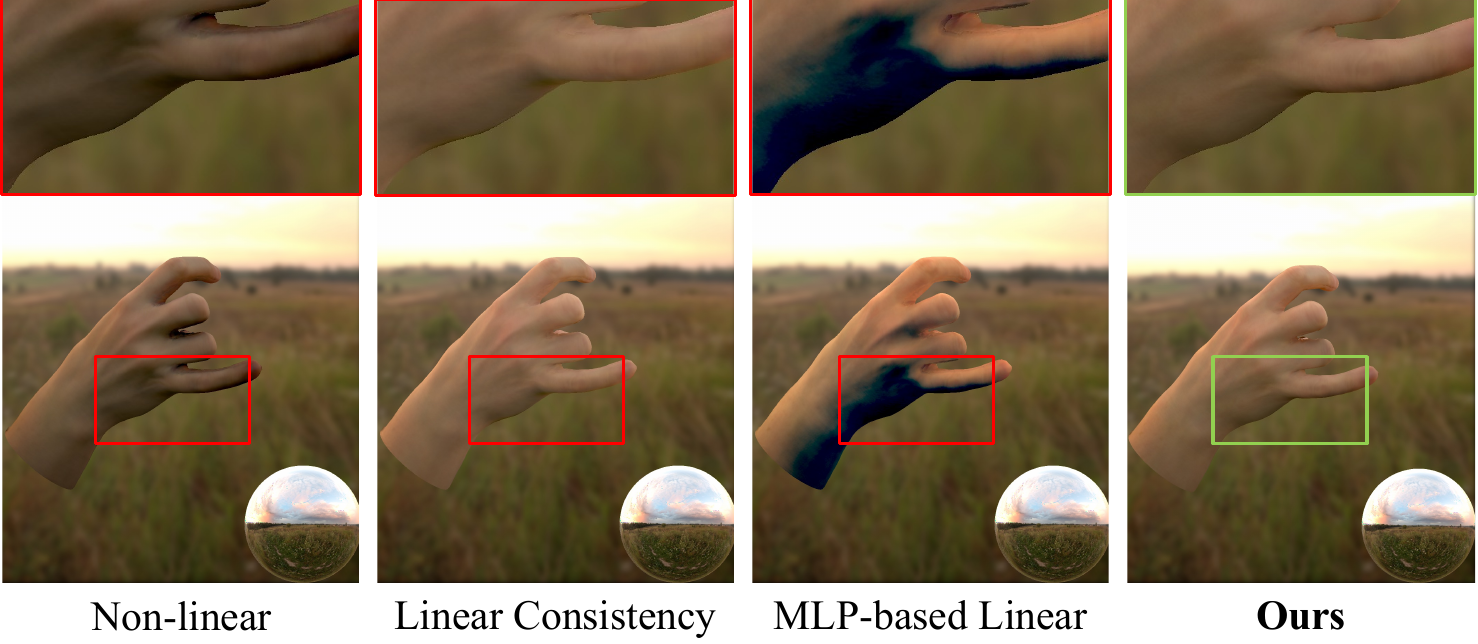}
    \caption{\textbf{Ablation study on the design of linear lighting model.} Our spatially varying linear lighting model produces realistic renderings, while the baseline methods fail to correctly model shadows or tend to be over smooth.}
    \label{fig:abl-llm}
    \vspace{-0.1in}
\end{figure}

\noindent \textbf{Ablation on Linear Lighting Model.} 
We investigate different designs of the linear lighting model. We compare our spatially varying linear lighting model with three alternatives: \textbf{1)} Non-linear model, where all non-linear activations are turned on; \textbf{2)} Non-linear model with a linearity consistency loss to constrain the linearity (Eq.~\ref{eq:linearity}) between output and input of the network; \textbf{3)} MLP-based linear model proposed in~\cite{yang2023towards}, where the lighting is represented as environment map and flattened to a 1-D vector.

Table~\ref{fig:abl-llm} shows that the proposed spatially varying linear lighting model achieves the best performance on both trained and unseen identities. Moreover, we also quantitatively evaluate on fully lit frames to validate the generalization to novel illumination. The non-linear baseline fails to generalize, which shows the importance of linearity in our network. The MLP-based linear model fails to correctly model the light transport due to the lack of pose-aware visibility change. Figure~\ref{fig:abl-llm} shows that the proposed lighting model produces the most realistic renderings, while the baseline methods struggle with correctly modeling shadows and detailed shading effects.

\begin{table}[t]
    \centering
    \begin{footnotesize}
    \caption{\textbf{Ablation studies of the lighting features on sequences with grouped lights}. The top three techniques are highlighted in \textcolor{rred}{red}, \textcolor{oorange}{orange}, and \textcolor{yyellow}{yellow}, respectively.}
    \label{tab:abl-feature}
    \setlength{\tabcolsep}{2mm}{
    \renewcommand\arraystretch{0.975}
    \resizebox{1.0\columnwidth}{!}{
    \begin{tabular}{c|cccccc} 
    \toprule
    \multirow{2}{*}{Method}& \multicolumn{3}{c}{\textbf{Trained Subject}}  & \multicolumn{3}{c}{\textbf{Unseen Subject}}\\ 
    & \footnotesize{PSNR\;$\uparrow$} & \footnotesize{SSIM\;$\uparrow$} & \footnotesize{LPIPS\;$\downarrow$} & \footnotesize{PSNR\;$\uparrow$} & \footnotesize{SSIM\;$\uparrow$} & \footnotesize{LPIPS\;$\downarrow$}\\  
    \midrule
    Phong&\cellcolor{oorange}25.53&\cellcolor{oorange}0.9257&\cellcolor{oorange}0.1382&\cellcolor{oorange}23.87&\cellcolor{rred}0.9148&\cellcolor{oorange}0.1523\\
    w/o Specular&24.97&0.9160&0.1498&23.20&0.9075&0.1607\\
    w/o Visibility&\cellcolor{yyellow}25.44&\cellcolor{yyellow}0.9228&\cellcolor{yyellow}0.1426&\cellcolor{yyellow}23.75&\cellcolor{oorange}0.9133&\cellcolor{yyellow}0.1533\\
    w/o Refiner&24.82&0.9144&0.1437&22.80&0.8993&0.1620\\
    \textbf{Full Model}&\cellcolor{rred}26.01&\cellcolor{rred}0.9270&\cellcolor{rred}0.1336&\cellcolor{rred}24.70&\cellcolor{yyellow}0.9121&\cellcolor{rred}0.1520\\
    \bottomrule
    \end{tabular}}}
    \end{footnotesize}
    \vspace{-0.15in}
\end{table}

\begin{figure}
    \centering
    \includegraphics[width=1.0\linewidth]{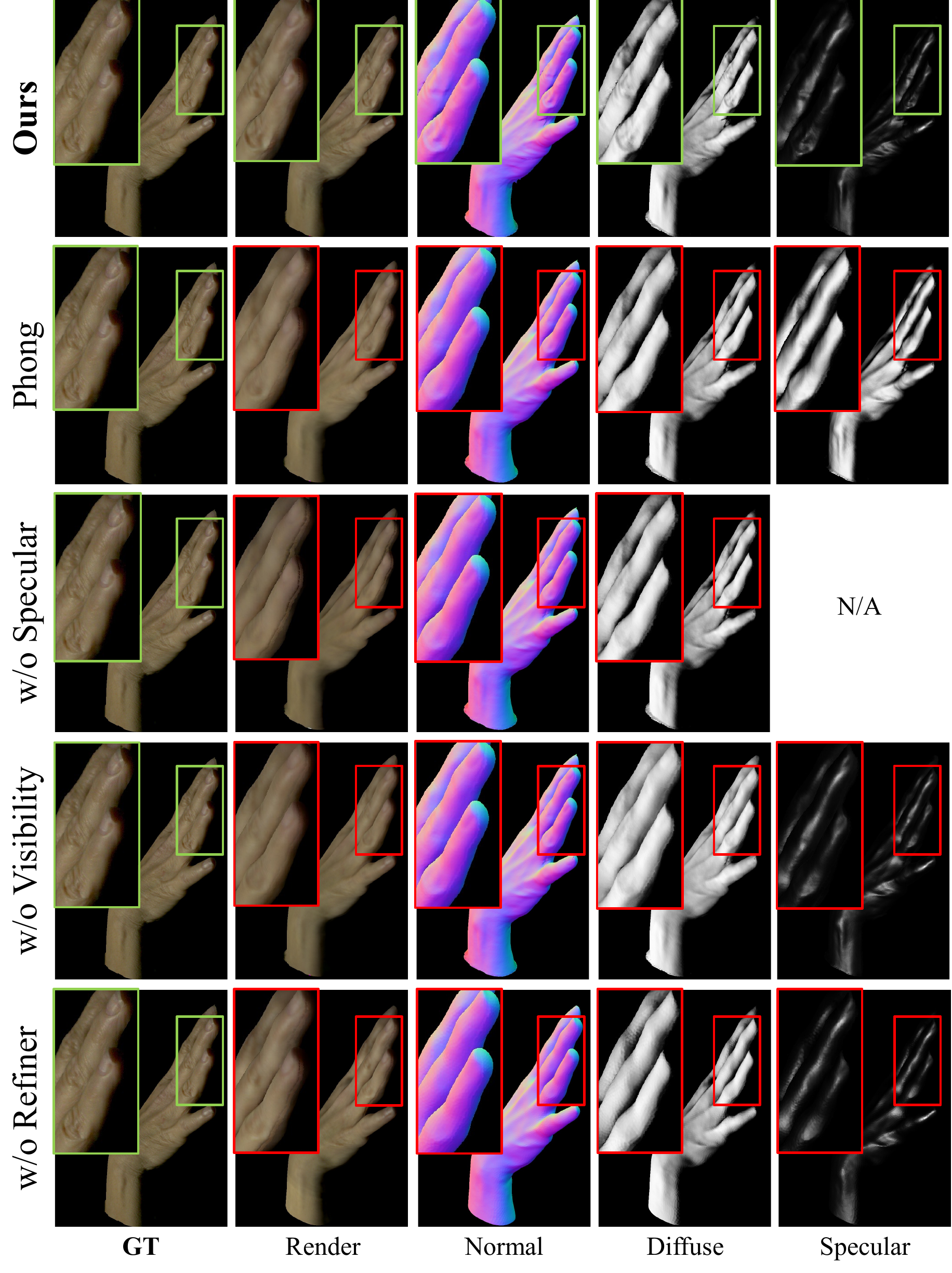}
    \vspace{-0.15in}
    \caption{\textbf{Ablation studies on the impact of lighting features and geometry refinement.} Notably, our full model can produce fine-grained geometry like wrinkles and nails as well as specular highlights (\eg little finger).}
    \label{fig:abl-feature}
    \vspace{-0.15in}
\end{figure}

\noindent \textbf{Ablation on Different Lighting Features.} 
We also evaluate the effectiveness of our lighting feature representation, \ie the diffuse $\mathbf{C}^d_{\mathrm{pb}}$ and specular feature $\mathbf{C}^s_{\mathrm{pb}}$. The contribution in the lighting features can be factored into \textbf{1)} the underlying BRDF, \textbf{2)} the visibility, and \textbf{3)} the geometry. As shown in Figure~\ref{fig:abl-feature}, our model reproduces specular highlights due to the more accurate specular feature from the optimized Disney BRDF. This indicates that the quality of the lighting feature determines the quality of the neural relighting. In fact, the neural relighting based on Phong specular features~\cite{totalrelighting,Iwase_2023} fails to produce accurate specular highlights. The visibility is also important for novel pose generalization. Compared with the model trained without visibility, our full model reproduces better soft shadows as well as detailed geometry. These observations are also supported by our quantitative evaluation presented in Table~\ref{tab:abl-feature}.

\noindent \textbf{Effectiveness of Geometry Refinement.} Our proposed neural-physical rendering offers the ability to refine geometry. As shown in Figure~\ref{fig:abl-feature}, our full model can produce more fine-grained details such as wrinkles and nails. The geometry refinement evidently prevents the neural renderer from hallucinating, leading to better generalization to novel identities and illuminations. The quantitative results in Table~\ref{tab:abl-feature} also support this observation.

\begin{table}[t]
    \centering
    \vspace{-0.1in}
    \begin{footnotesize}
    \caption{\textbf{Ablation studies of the proposed training objectives on sequences with grouped lights}. The top three techniques are highlighted in \textcolor{rred}{red}, \textcolor{oorange}{orange}, and \textcolor{yyellow}{yellow}, respectively.}
    \label{tab:abl-loss}
    \setlength{\tabcolsep}{2mm}{
    \renewcommand\arraystretch{0.975}
    \resizebox{1.0\columnwidth}{!}{
    \begin{tabular}{c|cccccc} 
    \toprule
    \multirow{2}{*}{Method}& \multicolumn{3}{c}{\textbf{Trained Subject}}  & \multicolumn{3}{c}{\textbf{Unseen Subject}}\\ 
    & \footnotesize{PSNR\;$\uparrow$} & \footnotesize{SSIM\;$\uparrow$} & \footnotesize{LPIPS\;$\downarrow$} & \footnotesize{PSNR\;$\uparrow$} & \footnotesize{SSIM\;$\uparrow$} & \footnotesize{LPIPS\;$\downarrow$}\\  
    \midrule
    w/o $\gL_{\mathrm{GAN}}$&\cellcolor{oorange}25.95&\cellcolor{rred}0.9271&0.1392&\cellcolor{yyellow}23.70&0.9113&0.1542\\
    w/o Light-aware $\gL_{\mathrm{GAN}}$&25.38&0.9215&\cellcolor{yyellow}0.1391&23.88&\cellcolor{oorange}0.9142&\cellcolor{yyellow}0.1521\\
    w/o L1 Reg $\gL_{\mathrm{reg}}$&\cellcolor{yyellow}25.52&\cellcolor{yyellow}0.9242&\cellcolor{oorange}0.1351&\cellcolor{oorange}23.90&\cellcolor{rred}0.9149&\cellcolor{rred}0.1491\\
    \textbf{Full Model}&\cellcolor{rred}26.01&\cellcolor{oorange}0.9270&\cellcolor{rred}0.1336&\cellcolor{rred}24.70&\cellcolor{yyellow}0.9121&\cellcolor{oorange}0.1520\\
    \bottomrule
    \end{tabular}}}
    \end{footnotesize}
    \vspace{-0.1in}
\end{table}

\begin{figure}
    \centering
    \includegraphics[width=1.0\linewidth]{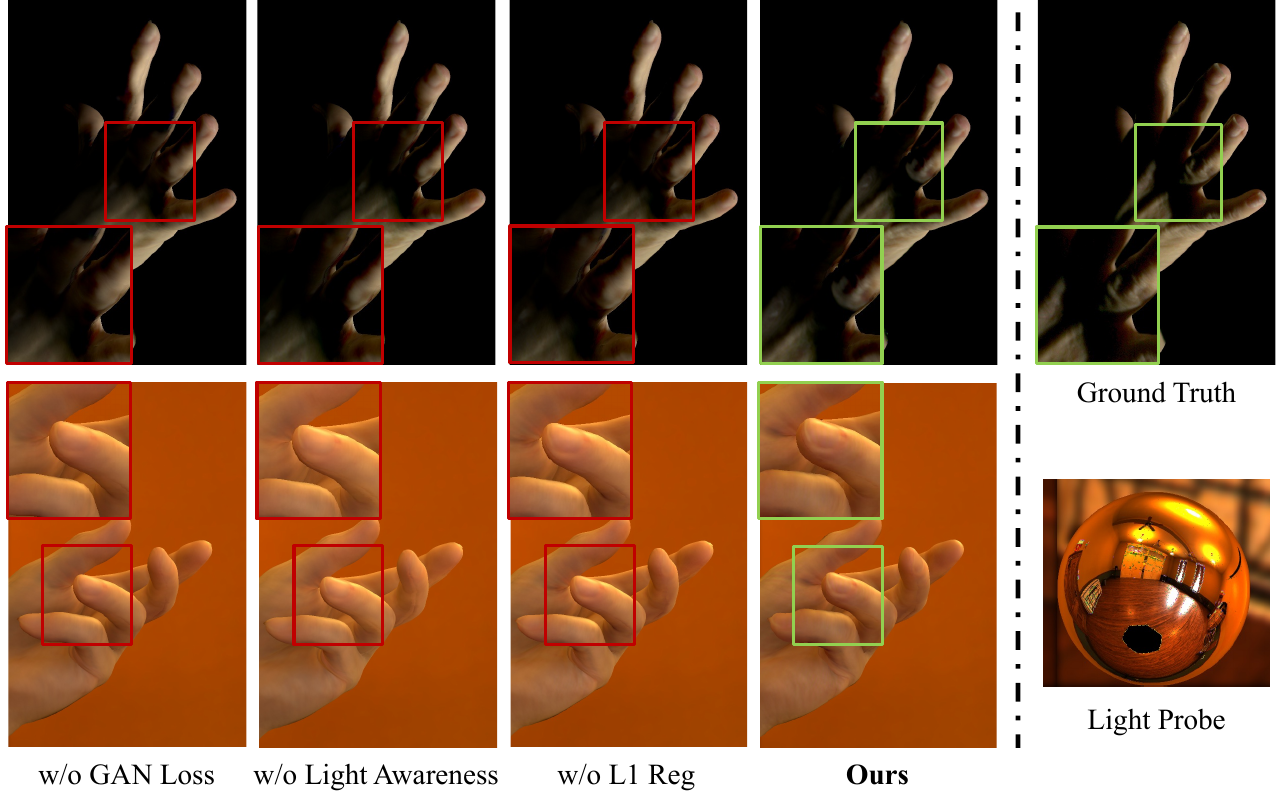}
    \caption{\textbf{Ablation studies on the proposed training objectives.} The adversarial loss improves the overall quality while the light awareness of the discriminator is critical for correct shadows. }
    \label{fig:abl-loss}
    \vspace{-0.1in}
\end{figure}

\noindent \textbf{Effectiveness of Adversarial Loss.}
We evaluate the effectiveness of the proposed lighting-aware adversarial loss. Quantitative results in Table~\ref{tab:abl-loss} suggest that this adversarial loss improves the overall quality. The lighting-aware discriminator further improves the fidelity. Figure~\ref{fig:abl-loss} validates that the adversarial loss is critical for reproducing shadow and detailed geometry. It also enhances specular highlights.

\noindent \textbf{Effectiveness of L1 Regularization.}
We validate the effectiveness of the L1 regularization on the intermediate features of the linear lighting model. Table~\ref{tab:abl-loss} and Figure~\ref{fig:abl-loss} illustrate that, compared with the model without L1 regularization, our model achieves better modeling of the soft shadows and appearance. Its effectiveness is more evident in the temporal sequence with less flickering artifacts. Please refer to our supplemental video.

\subsection{Quick Adaptation on Unseen Subjects}
\label{sec:instant}

\begin{figure}
    \centering
    \vspace{-0.1in}
    \includegraphics[width=1.0\linewidth]{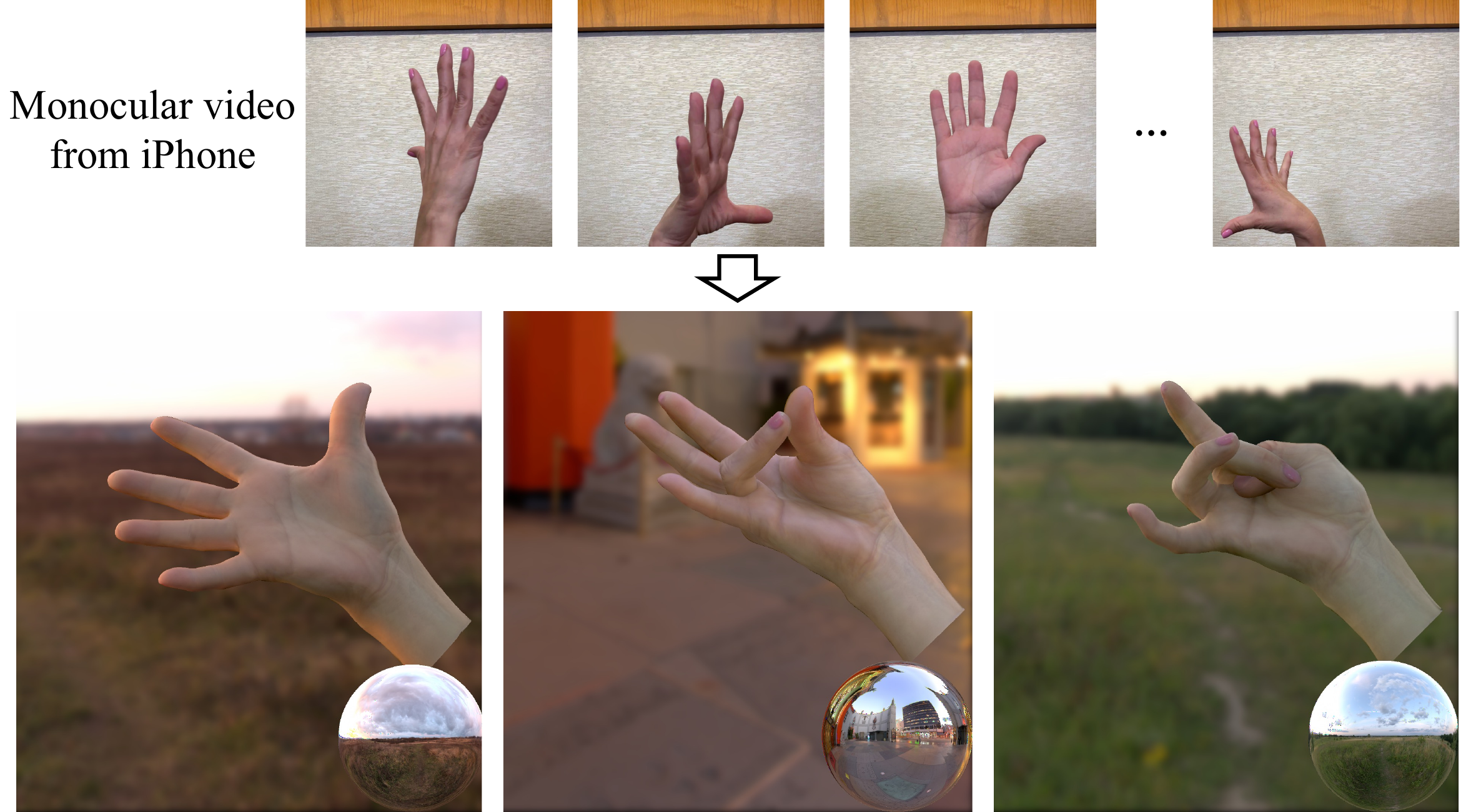}
    \caption{\textbf{Quick Personalization of \nickname\ from iPhone captures.} Given a casual phone scan, we fit hand geometry (Sec.~\ref{sec:uhm}), unwrap RGB images to get the mean texture, and feed into \nickname. Our model instantly enables photorealistic relighting in any poses and illuminations without finetuning.}
    \label{fig:instant}
    \vspace{-0.1in}
\end{figure}

As shown in Figure~\ref{fig:instant}, we demonstrate the quick personalization of \nickname\ to get a relightable and animatable hand from a monocular iPhone video. To achieve this, we fit a template hand model~(Sec.~\ref{sec:uhm}) to input images by optimizing pose parameters and the identity latent code based on foreground segmentation and 2D keypoints following a similar pipeline to~\cite{karunratanakul2023harp}. Then we unwrap input RGB images based on the fitted geometry to obtain the mean texture $\gT$, and refine it to remove shadows~\cite{karunratanakul2023harp}. Finally, we directly feed the mean texture and the coarse mesh into \nickname\ to render it with any poses and environment maps. Note that while \nickname\ can instantly create a personalized relightable hand model without any finetuning, the aforementioned preprocessing stage takes several hours.

\section{Conclusion}
\label{sec:discussion}
We have introduced URHand, the first universal relightable hand model that generalizes across viewpoints, poses, illuminations, and identities. We show that scalable cross-identity training for high-fidelity relightable hands now is possible with our physics-inspired spatially varying linear lighting model and hybrid neural-physical learning framework. Our experiments indicate that \nickname\ even generalizes beyond studio data by showing quick personalization from a phone scan. 

\noindent \textbf{Limitation and Future Works.}
Since we learn global light transport with far-field lighting, it does not guarantee correct light transport with near-field lighting. Nevertheless, our work achieves plausible near-field relighting similarly to~\cite{bi2021deep,Iwase_2023,sarkar2023litnerf,xu2023renerf}.
Currently the quick personalization requires the complete mean texture of a target hand. Thus, it does not work with a single image. One future work would be inpainting the texture from a single image to enable single-view relightable hand reconstruction. As our hand model is only driven by hand poses, it cannot capture appearance variations due to blood pressure or temperature changes. As recently demonstrated in~\cite{moon2023dataset}, photorealistic relightable hands can be used to augment training data for image-based pose regression tasks. Using \nickname\ to synthesize large-scale two-hand or hand-to-object interaction images with diverse identities is also fruitful.


{
    \small
    \bibliographystyle{ieeenat_fullname}
    \bibliography{main}
}

\clearpage
\setcounter{page}{1}
\maketitlesupplementary

\section{Demo Video and Additional Results}
\label{sec:demo}
We provide the supplementary video on our project page (\url{https://frozenburning.github.io/projects/urhand}), which includes more visual results and additional discussions of our work. Specifically, it contains:
\begin{itemize}
    \item Motivation and key features of \nickname.
    \item An animated overview and illustration of the proposed framework.
    \item Video comparisons with baseline methods.
    \item Additional qualitative results with diverse identities, including \textbf{1)} relighting with monochrome directional light, \textbf{2)} relighting with arbitrary environment map, and \textbf{3)} quick personalization from a phone scan with corresponding relighting results with environment maps.
\end{itemize}

\section{Network Architecture}
\label{sec:network-arch}
In this section, we provide the details of our network architecture and hyperparameters for our hand geometry model (Sec.~\ref{sec:uhm-arch}), physical branch (Sec.~\ref{sec:phys-arch}), and neural branch (Sec.~\ref{sec:neural-arch}), respectively.

\begin{figure*}[t]
    \centering
    \includegraphics[width=1.0\linewidth]{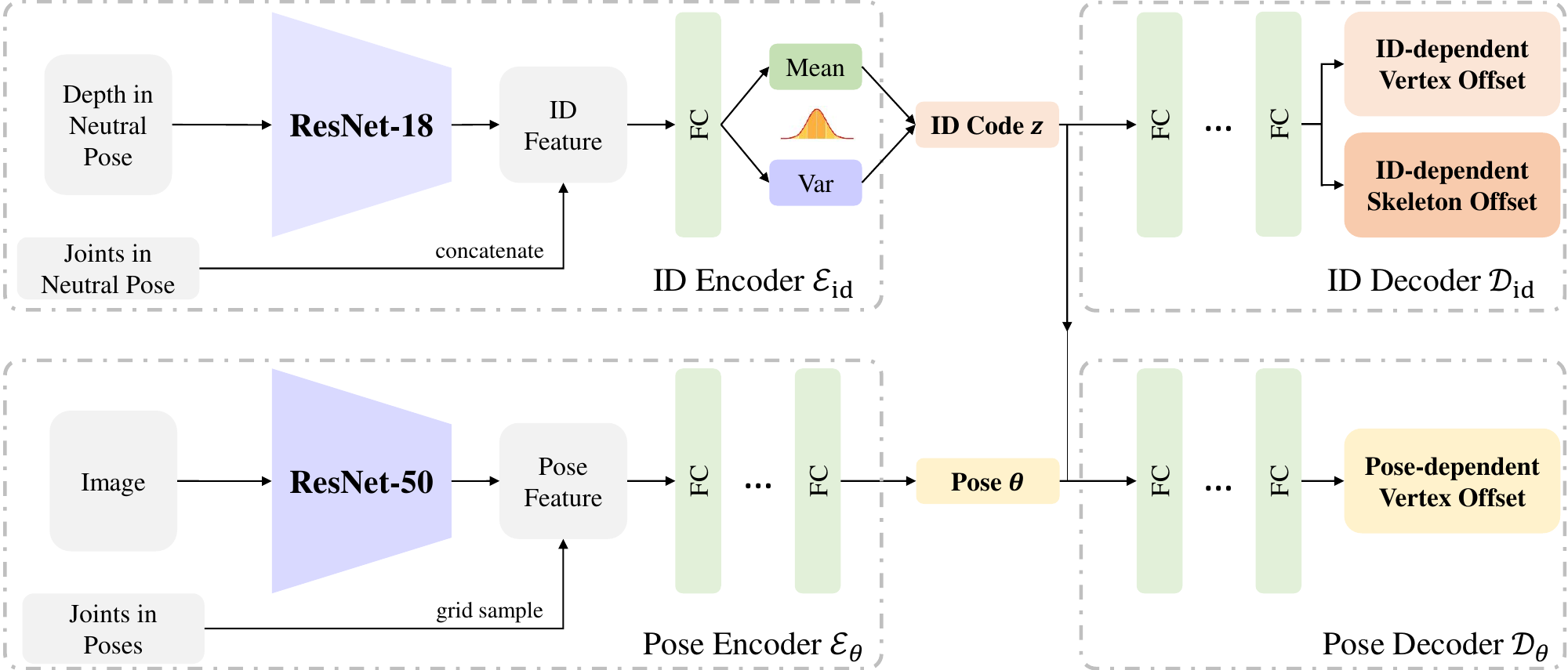}
    \caption{\textbf{The architecture of our hand geometry autoencoder.} The identity encoder $\gE_{\mathrm{id}}$ takes as input the depth map in the neutral pose and the coordinates of joints in the neutral pose, which predicts the mean and variance of the distribution of the identity code (\ie ID code). The identity decoder $\gD_{\mathrm{id}}$ learns to decode the identity-dependent offset of vertices and skeletons from the ID code $z$. The pose encoder $\gE_{\theta}$ directly regresses hand pose $\theta$ from the input image and the coordinates of joints. The pose decoder $\gD_{\theta}$ learns to predict the pose-dependent offset of vertices given the pose $\theta$ and ID code $z$.}
    \label{fig:uhm-arch}
\end{figure*}

\subsection{Hand Geometry Autoencoder}
\label{sec:uhm-arch}
We design an autoencoder to obtain accurate hand tracking and geometry from input fully lit frames similar to~\cite{moon2020deephandmesh}. The architecture of this autoencoder $\{\gE_{\mathrm{id}}, \gD_{\mathrm{id}}, \gE_{\theta}, \gD_{\theta}\}$ is illustrated in Figure~\ref{fig:uhm-arch}. Specifically, it consists of an identity encoder $\gE_{\mathrm{id}}$, an identity decoder $\gD_{\mathrm{id}}$, a pose encoder $\gE_{\theta}$, and a pose decoder $\gD_{\theta}$. 

The identity encoder $\gE_{\mathrm{id}}$ takes as input the depth map in the neutral pose and the coordinates of joints in the neutral pose, which predicts the mean and variance of the distribution of the identity code (\ie ID code). The inputs of the identity encoder have normalized viewpoints by rigidly aligning them to a reference coordinate system; hence both pose and viewpoints are normalized and only identity information is included in them. The identity decoder $\gD_{\mathrm{id}}$ learns to decode the identity-dependent offset of joints and vertices from the ID code $z$. The identity-dependent offset of joints is responsible for adjusting 3D joint coordinates in the template space for each identity, and the offset of vertices are for adjusting 3D vertices in the template space for each identity. 
The pose encoder $\gE_{\theta}$ directly regresses hand pose $\theta$ from the input image and the coordinates of joints. The pose decoder $\gD_{\theta}$ learns to predict the pose-dependent offset of vertices given the pose $\theta$ and ID code $z$. 
To get posed 3D meshes, we apply the three types of correctives to the template mesh and perform linear blend skinning with the estimated 3D pose from the pose encoder. 

We train this autoencoder $\{\gE_{\theta}, \gD_{\theta}, \gE_{\mathrm{id}}, \gD_{\mathrm{id}}\}$ on fully lit frames with all identities to obtain a general hand tracker. The autoencoder is trained by minimizing 1) $L1$ distance between joint coordinates, 2) point-to-point $L1$ distance from 3D scans with weight 10, 3) KL-divergence of the ID code with weight 0.001, and 4) various regularizers like Moon~\etal~\cite{moon2020deephandmesh}. We freeze it during the training of \nickname\ as well as the quick personalization from phone scans.

\subsection{Physical Branch}
\label{sec:phys-arch}
The physical branch of \nickname\ consists of a 2D U-Net~\cite{ronneberger2015u} $\gF_{\mathrm{G}}$ and a parametric BRDF~\cite{disneybrdf} $\gF_{\mathrm{pb}}$, where only the U-Net $\gF_{\mathrm{G}}$ contains optimizable parameters. The U-Net encoder is a 6-layer convolutional neural network (CNN) with channel sizes $(3, 64, 64, 64, 64, 64, 64)$, which takes as input the mean texture $\gT \in \mathbb{R}^{1024\times 1024 \times 3}$. Here, the hand pose $\theta$ is tiled into a UV-aligned 2D feature map $\theta'$, concatenated with the output feature from the U-Net encoder as a joint feature $\mathbf{F}_{\theta, \mathrm{id}}$, and passed to the U-Net decoder. The U-Net decoder is a 6-layer CNN with skip connections from the U-Net encoder, with channel sizes $(64, 64, 64, 64, 64, 64, 2)$. We use a transposed convolution layer followed by bilinear interpolation as the upsampling layer in the U-Net decoder. The U-Net decoder predicts the displacement map $\delta d \in \mathbb{R}^{1024\times1024}$ and the roughness map $\beta \in \mathbb{R}^{1024\times 1024}$. We unwrap the coarse mesh $\gM$ from our hand geometry autoencoder into the UV space to obtain the corresponding coarse normal map $\mathbf{n} \in \mathbb{R}^{1024\times 1024 \times 3}$. The predicted displacement map is applied on top of this coarse normal map to obtain the refined normal map $\hat{\mathbf{n}}$ according to Eq.~\ref{eq:displacement} in the main paper.

The parametric BRDF $\gF_{\mathrm{pb}}$ takes as input the refined normal map $\hat{\mathbf{n}}$, the roughness map $\beta$, light $\gL = \{L_i(\mathbf{\omega}_i)\}_i$, and view direction $\mathbf{d}$. The physics-inspired shading features $\mathbf{F}_{\mathrm{pb}} = \{\mathbf{C^d_{\mathrm{pb}}, \mathbf{C}^s_{\mathrm{pb}}}\}$ are computed accordingly. Specifically, the diffuse feature $\mathbf{C^d_{\mathrm{pb}}}$ is computed as:
\begin{equation}
\label{eq:phys-diff-feature}
    \mathbf{C^d_{\mathrm{pb}}} = \int L_i(\mathbf{\omega}_i) \cdot \mathbf{V}_i \cdot (\mathbf{\omega}_i \cdot \hat{\mathbf{n}}) d\mathbf{\omega}_i,
\end{equation}
where $L_i(\mathbf{\omega}_i)$ is the light intensity from the incident direction $\mathbf{\omega}_i$, $\mathbf{V}_i$ is the visibility given the light $L_i$. Furthermore, the specular feature $\mathbf{C}^s_{\mathrm{pb}}$ is computed as:
\begin{align}
\label{eq:phys-spec-feature}
    \mathbf{C}^s_{\mathrm{pb}} &= \int {D\cdot F\cdot G} \cdot L_i(\mathbf{\omega}_i) \cdot \mathbf{V}_i \cdot (\mathbf{\omega}_i \cdot \hat{\mathbf{n}}) d\mathbf{\omega}_i,\\
    D &= \frac{\beta^4}{\pi [(\mathbf{h} \cdot \hat{\mathbf{n}})^2 (\beta^4 - 1) + 1]^2},\\
    F &= F_0 + (1 - F_0) \cdot 2^{[\lambda_{F1}(\mathbf{d}\cdot \mathbf{h}) + \lambda_{F2}](\mathbf{d} \cdot \mathbf{h})},\\
    G &= \frac{1}{4 [(\hat{\mathbf{n}} \cdot \mathbf{d}) (1-K) + K] [(\hat{\mathbf{n}} \cdot \mathbf{\omega}_i) (1-K) + K]},\\
    \mathbf{h} &= \frac{\mathbf{\omega}_i + \mathbf{d}}{||\mathbf{\omega}_i + \mathbf{d}||}, \;\; K = \frac{(\beta + 1)^2}{8},
\end{align}
where we set Fresnel coefficient $F_0 = 0.04$, $\lambda_{F1} = -5.55473$, and $\lambda_{F2} = -6.98316$, respectively.

\begin{figure}
    \centering
    \includegraphics[width=1.0\linewidth]{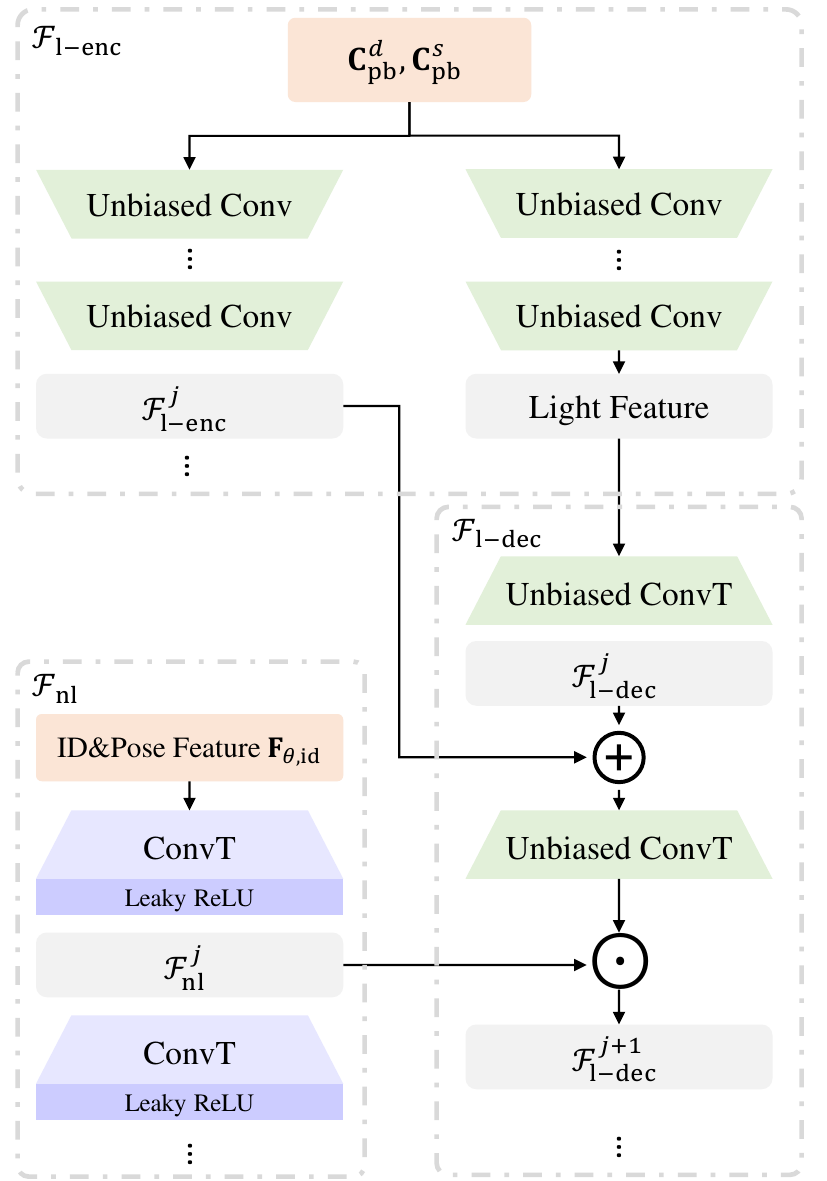} 
    \caption{\textbf{The architecture of the neural branch of \nickname.} The neural branch of our model consists of a non-linear network $\gF_{\mathrm{nl}}$ and a linear network $\gF_{\mathrm{l}}$ (\ie linear lighting model). Notably, we remove all non-linear activation and use the convolutional layer without bias in the linear network so that the linearity of the output is explicitly kept \wrt the input physics-inspired shading feature $\{\mathbf{C}^d_{\mathrm{pb}}, \mathbf{C}^s_{\mathrm{pb}}\}$. This figure also illustrates how Eq.~\ref{eq:gb-injection} in the main paper is implemented within our network.}
    \label{fig:neural-arch}
\end{figure}

\subsection{Neural Branch}
\label{sec:neural-arch}
The neural branch of \nickname\ consists of a non-linear network $\gF_{\mathrm{nl}}$ and a linear network $\gF_{\mathrm{l}}$ (\ie linear lighting model). We illustrate the detailed architecture of the neural branch in Figure~\ref{fig:neural-arch}. Specifically, the non-linear network $\gF_{\mathrm{nl}}$ is a 7-layer CNN with channel sizes $(128, 256, 128, 128, 64, 32, 16, 4)$, which takes as input the pose- and identity-dependent joint feature $\mathbf{F}_{\theta, \mathrm{id}}$. The linear network $\gF_{\mathrm{l}}$, namely the linear lighting model, consists of an encoder $\gF_{\mathrm{l-enc}}$ and a decoder $\gF_{\mathrm{l-dec}}$. The linear encoder $\gF_{\mathrm{enc}}$ consists of unbiased convolutional layers which takes as input the concatenated physics-inspired shading features $\{\mathbf{C}^d_{\mathrm{pb}}, \mathbf{C}^s_{\mathrm{pb}}\}$. The linear decoder is a 7-layer unbiased CNN with channel sizes $(128, 256, 128, 128, 64, 32, 16, 4)$. We fuse the linear features from $\gF_{\mathrm{l-enc}}$ and the non-linear features from $\gF_{\mathrm{nl}}$ as layer-wise modulation at each layer of the linear decoder $\gF_{\mathrm{l-dec}}$ according to Eq.~\ref{eq:gb-injection} in the main paper. The predicted gain map $g \in \mathbb{R}^{1024\times 1024 \times 3}$ and bias map $b \in \mathbb{R}^{1024\times 1024}$ contributes to the final rendering according to Eq.~\ref{eq:gb-rendering} in the main paper.

\section{Adaptation to a Phone Scan}
In this section, we present the details of how to quickly adapt \nickname\ to a personalized use case from a phone scan. We use a single iPhone 12 to scan a hand, which incorporates a depth sensor that can be used to extract better geometry of the user’s hand.
Our phone scans include hands with neutral finger poses and static 3D global translations with varying 3D global rotations to expose most of the hand surfaces.

We pre-process the phone scan with 1) our in-house 2D hand keypoint detector to obtain 2D hand joint coordinates and 2) RVM~\cite{lin2022robust} to obtain the foreground mask of the phone scan.
After the preprocessing, we optimize 3D global rotation, 3D pose, 3D global translation, and ID code of the phone scan.
The 3D global rotation, 3D pose, and 3D global translation are optimized for each frame, and a single ID code is shared across all frames as all frames are from a single identity.
We optimize them by minimizing 1) $L1$ distance between projected 2D joint coordinates and targets 2) $L1$ distance between differentiably rendered masks and targets with weight 50, and 3) $L1$ distance between differentiably rendered depth maps and targets with weight 100.
After the optimization, we unwrap per-frame images to UV space and average intensity values at each texel considering the visibility to get the unwrapped texture map.

To remove shadows from the unwrapped textures, we first obtain the average color of the foreground pixels of captured images.
Then, we optimize shadow as a 1-channel difference (\textit{i.e.}, darkness difference) between the averaged color and the captured image in the UV space.
To prevent the shadow from dominating local sharp textures (\textit{i.e.}, hairs and tattoos), we apply a total variation regularizer to the shadow.
The unwrapped texture without the shadow is simply obtained by dividing the unwrapped texture by the shadow.
We empirically observed that such a statistical approach produces better shadow than the physics-based approach of HARP~\cite{karunratanakul2023harp}, which assumes a single point light, as there is often more than one light source in the scan environment. Then, we take this texture map after shadow removal as the input to \nickname\ for relighting without any finetuning.

\section{Implementation of Baselines}
\label{sec:baseline-impl}
In this section, we present implementation details of our baseline methods for comparisons in the main paper. Specifically, we introduce our modifications to RelightableHands~\cite{Iwase_2023} and Handy~\cite{potamias2023handy} in Sec.~\ref{sec:sota-impl}. Moreover, we introduce our implementations of all baselines for ablation studies in Sec.~\ref{sec:abl-impl}. 

\subsection{Methods for Main Comparisons}
\label{sec:sota-impl}
\noindent \textbf{RelightableHands}~\cite{Iwase_2023} is originally proposed for per-identity relightable appearance reconstruction tailored with volumetric representation~\cite{lombardi2021mixture}. For a fair comparison, we re-implement it with our mesh-based representation. Specifically, we leverage a U-Net as the texture decoder $\gA_{\mathrm{tex}}$ that takes as input the UV-aligned view direction, light direction, and visibility. The pose parameter is tiled into a UV-aligned feature map and concatenated with the bottleneck representation of the U-Net. This texture decoder $\gA_{\mathrm{tex}}$ predicts texture map $\mathbf{T} \in \mathbb{R}^{1024\times 1024 \times 3}$ and shadow map $\mathbf{S} \in \mathbb{R}^{1024\times 1024}$ in the UV space. The final texture $\mathbf{C}$ for rendering is obtained as:
\begin{equation}
    \mathbf{C} = \sigma(\mathbf{S})(\mathrm{ReLU}(\lambda_s \mathbf{T}) +\lambda_b),
\end{equation}
where $\sigma(\cdot)$ is the sigmoid function, $\mathrm{ReLU}(x) = \max(0, x)$, $\lambda_s$ is a scale factor, and $\lambda_b$ is a bias parameter. In our experiments, we set $\lambda_s = 25$, and $\lambda_s = 100$, respectively.

\noindent \textbf{Handy}~\cite{potamias2023handy} leverages the parametric hand model~\cite{romero2017mano} as the shape representation and StyleGAN3~\cite{Karras2021} as the texture model. As the shape and latent code regressor are not publicly available, we cannot infer the latent code $w$ or shape parameters from input images as in the original paper. Instead, we fit hand shape parameters using our multiview fully lit frames on the training segments. Then, we do the StyleGAN inversion following~\cite{roich2021pivotal}. Specifically, we randomly initialize the latent code $w$ and optimize it given the reconstruction loss between the ground truth fully lit images and the images rendered with the current predicted texture map. In our experiments, we optimize 50,000 iterations for each identity. We use the Adam~\cite{kingma2014adam} optimizer with the initial learning rate as $1\times 10^{-3}$. Once the inversion is done, we take the latent code $w$ and feed it into the pretrained texture model of Handy to get the unwrapped texture map. We treat the unwrapped texture map as the albedo map for physically based relighting evaluation.

\subsection{Baselines for Ablation Studies}
\label{sec:abl-impl}
\noindent \textbf{Non-linear model} is based on our full model but we add $\mathrm{LeakyReLU}$ function to all layers in the original linear network $\gF_{\mathrm{l}}$ which breaks the linearity.

\noindent \textbf{Linear consistency model} is based on the aforementioned non-linear model. We additionally constrain the linearity of this non-linear network by applying linearity consistency loss during training. Specifically, for every $n$ iterations, we augment two physics-inspired shading features with two random scalars, \ie $a_1 \mathbf{F}^1_{\mathrm{pb}} + a_2 \mathbf{F}^2_{\mathrm{pb}},$ where $a_1, a_2 \in (0, 1)$. The linearity consistency loss is defined as:
\begin{equation}
\small
    \gL_{\mathrm{lc}} = ||a_1\gF_{\mathrm{l}}(\mathbf{F}^1_{\mathrm{pb}})+a_2\gF_{\mathrm{l}}(\mathbf{F}^2_{\mathrm{pb}}) - \gF_{\mathrm{l}}(a_1 \mathbf{F}^1_{\mathrm{pb}} + a_2 \mathbf{F}^2_{\mathrm{pb}})||_2
\end{equation}

\noindent \textbf{MLP-based linear model}~\cite{yang2023towards} is a variant of the linear lighting model with no spatially varying lighting feature. We replace the encoder of linear network $\gF_{\mathrm{l-enc}}$ as a one-layer MLP without bias. It takes as input the environment map with a resolution of $3\times 16\times 32$, and predicts the lighting feature. Then we reshape the prediction into a UV-aligned feature map with a resolution of $128\times 16\times 16$ and feed into the decoder of linear network to predict the final gain and bias map for neural rendering. 

\noindent \textbf{Phong} based model is implemented by replacing our physics-inspired shading feature $\mathbf{F}_{\mathrm{pb}} = \{\mathbf{C}^d_{\mathrm{pb}}, \mathbf{C}^s_{\mathrm{pb}}\}$ with simple diffuse and specular feature from the Phong reflectance model. This neural relighting model is similar to~\cite{Iwase_2023, bi2021deep} with no learnable material parameter.

\noindent \textbf{w/o Specular} is the baseline where we dropout the specular feature $\mathbf{C}^s_{\mathrm{pb}}$ during training.

\noindent \textbf{w/o Visibility} is the baseline where we do not incorporate visibility $\mathbf{V}_i$ when compute the physics-inspired shading feature in Eq.~\ref{eq:phys-diff-feature} and Eq.~\ref{eq:phys-spec-feature}.

\noindent \textbf{w/o Refiner} is the baseline where we only use the normal map $\mathbf{n}$ from the coarse geometry without further refinement during training.

\noindent \textbf{w/o $\gL_{\mathrm{GAN}}$} is the baseline trained with the reconstruction loss $\gL_{\mathrm{img}}$ and L1 regularization $\gL_{\mathrm{reg}}$ only.

\noindent \textbf{w/o Light-aware $\gL_{\mathrm{GAN}}$} is the baseline trained with the vanilla adversarial loss without conditional discriminator. Specifically, the adversarial loss of Eq.~\ref{eq:gan-loss} in the main paper degrades to $\gL_{\mathrm{GAN}} = \log \gF_{\mathrm{D}}(I) + \log [1 - \gF_{\mathrm{D}}(\hat{I})]$, where $I$ is the ground truth and $\hat{I}$ is the rendered image.

\noindent \textbf{w/o L1 Reg} is the baseline trained with the reconstruction loss $\gL_{\mathrm{img}}$ and lighting-aware adversarial loss $\gL_{\mathrm{GAN}}$ only.

\end{document}